\documentclass{article}

 \PassOptionsToPackage{numbers, compress}{natbib}
\usepackage[preprint]{neurips_2026}

\usepackage[utf8]{inputenc}
\usepackage[T1]{fontenc}
\usepackage{hyperref}
\usepackage{xurl}
\usepackage{booktabs}
\usepackage{amsfonts}
\usepackage{nicefrac}
\usepackage{microtype}
\usepackage{graphicx}        % ← 添加这一行
\usepackage[accsupp]{axessibility}
\usepackage{multirow}
\usepackage[table]{xcolor}   % 先加载 xcolor
\usepackage{arydshln}
\usepackage{pifont}
\usepackage{amsmath}
\usepackage{amssymb}
\usepackage{bussproofs}
\usepackage{indentfirst,float}
\usepackage{arydshln}
\usepackage{makecell}
\usepackage{array}           % simple URL typesetting
\usepackage{wrapfig}
\usepackage{tcolorbox}
\definecolor{boxbg}{HTML}{DAE3F3}
\newtcolorbox{conclusionbox}{
colback=boxbg!50,colframe=boxbg!50,boxrule=0pt,arc=6pt,auto outer arc,left=5pt,right=5pt,top=2pt,bottom=2pt
}
\usepackage{caption}

\definecolor{darkgreen}{RGB}{0, 100, 0}
\definecolor{cvprblue}{rgb}{0.21,0.49,0.74}
\definecolor{mred}{RGB}{238, 34, 12}
\definecolor{mgreen}{RGB}{1, 127, 0}
\definecolor{mblue}{RGB}{0, 77, 158}
\definecolor{mmblue}{RGB}{0, 77, 128}
\newcommand{\bred}[1]{\textbf{\textcolor{red}{#1}}}
\newcommand{\blue}[1]{\textbf{\textcolor{mmblue}{#1}}}
\hypersetup{
colorlinks=true,        % 如果想要彩色链接
linkcolor=blue,        % 用于普通链接的颜色
filecolor=magenta,     % 用于文件链接的颜色
urlcolor=magenta,          % 用于URL的颜色
}

\title{LL-Bench: Rethinking Low-Level Vision Evaluation in the Era of Large-Scale Generative Models}

\author{ Lu Liu$^{1}$, Huiyu Duan$^{1}$,  Chenxin Zhu$^{1}$, Jintong Lu$^{1}$, Haoyun Jiang$^{1}$,   \\[2pt] 
\textbf{Liu Yang} $^{1}$, \textbf{Qiang Hu}$^{1}$, \textbf{Guangtao Zhai}$^{1}$, \textbf{Xiaoyun Zhang}$^{1}$ \\[2pt]
$^{1}$Shanghai Jiao Tong University, Shanghai, China}
\begin{document}

\maketitle

\begin{abstract}
Large-scale generative models have demonstrated remarkable capabilities across image generation and editing tasks. However, their performance in low-level vision tasks, which require pixel-wise control, remains insufficiently studied. To address this gap, we introduce \textbf{LL-Bench}, a comprehensive \textbf{Benchmark} for evaluating the capabilities of large-scale generative models on \textbf{L}ow-\textbf{L}evel vision tasks. The benchmark comprises 2,469 real-world degraded images covering 16 low-level degradation tasks, and 28,919 restored images produced by 10 state-of-the-art large-scale generative models and 21 conventional restoration models, which are annotated with 152,020 expert-level pairwise human preferences and 28,334 quality scores. Built upon LL-Bench, we present a systematic diagnosis that reveals the performance boundaries and unique failure modes of large-scale generative models across diverse low-level vision tasks, compared with conventional representative restoration approaches. Moreover, we investigate the effectiveness of current quality evaluation metrics on LL-Bench, which exhibit significant discrepancy with human ratings. To better align restored-image quality assessment with human preferences, we further propose \textbf{LL-Score}, an MLLM-based evaluator that captures both restoration quality and hallucination existence. Extensive experiments demonstrate that LL-score not only outperforms existing image quality assessment metrics, but also serves as a promising reward model for training generative models on low-level vision tasks. The dataset and code will be publicly available at \url{https://github.com/MediaX-SJTU/LL-Bench}.

\end{abstract}

\section{Introduction}

Large-scale generative models (LGMs) such as NanoBananaPro~\cite{NanoBananaPro} and GPT-Image-2~\cite{GPT-Image-2.}, pre-trained on massive datasets with billions of parameters, have demonstrated remarkable capabilities and strong generalization across a wide range of image generation and editing tasks. 
These capabilities make them particularly promising for low-level vision tasks, including image restoration and enhancement, which naturally align with the image-to-image paradigm by transforming low-quality inputs into high-quality outputs. 
However, despite their impressive generative abilities, the capabilities and limitations of LGMs in low-level vision tasks remain insufficiently explored.

Unlike high-level generation and editing tasks, which focus on perceptual quality and semantic consistency, low-level vision tasks such as super-resolution, deblurring, enhancement, and deraining demand faithful pixel-level reconstruction and control.
Beyond producing visually pleasing outputs, models are required to preserve local textures, fine details, color distributions, and structural consistency with respect to the degraded input image.
Consequently, low-level restoration inherently involves a challenging trade-off between perceptual quality and fidelity, particularly for generative models originally trained for creative synthesis and instruction following.
Thus, evaluation introduces a significant challenge for low-level vision tasks in the era of large-scale generative models, since existing full-reference (FR) metrics, \textit{e.g.,} PSNR, SSIM~\cite{ssim}, LPIPS~\cite{lpips}, are largely developed for deterministic reconstruction and often fail to reflect perceptual quality in generative outputs, while no-reference (NR) metrics provide limited signals regarding fidelity to the input content.
A unified, human-aligned evaluation metric for generative low-level vision is also required.

To address this gap, we first propose \textbf{LL-Bench}, a comprehensive human-preference \textbf{benchmark} for \textbf{L}ow-\textbf{L}evel vision. Specifically, LL-Bench contains \textit{2,469 real-world degraded images} covering \textit{16} degradation types and diverse semantic categories. 
Based on these images, 28,919 restored images are generated by \textit{10} state-of-the-art large-scale generative models and \textit{21} representative specialist and all-in-one restoration methods. 
We further collect \textit{152,020 expert-level pairwise annotations} on overall restoration quality, yielding \textit{28,334} estimated Bradley-Terry (B-T) scores, and per-image hallucination judgments.       
Based on LL-Bench, we conduct a comprehensive analysis across models, tasks, and degradation categories, jointly comparing large-scale generative models with conventional restoration methods. Our analysis reveals clear capability boundaries and characteristic failure modes of generative models in low-level vision, highlighting where they remain competitive, where they lag behind specialist restoration models, and how their restoration behaviors differ under diverse degradation conditions.

\begin{figure}
\centering
\includegraphics[width=1\linewidth]{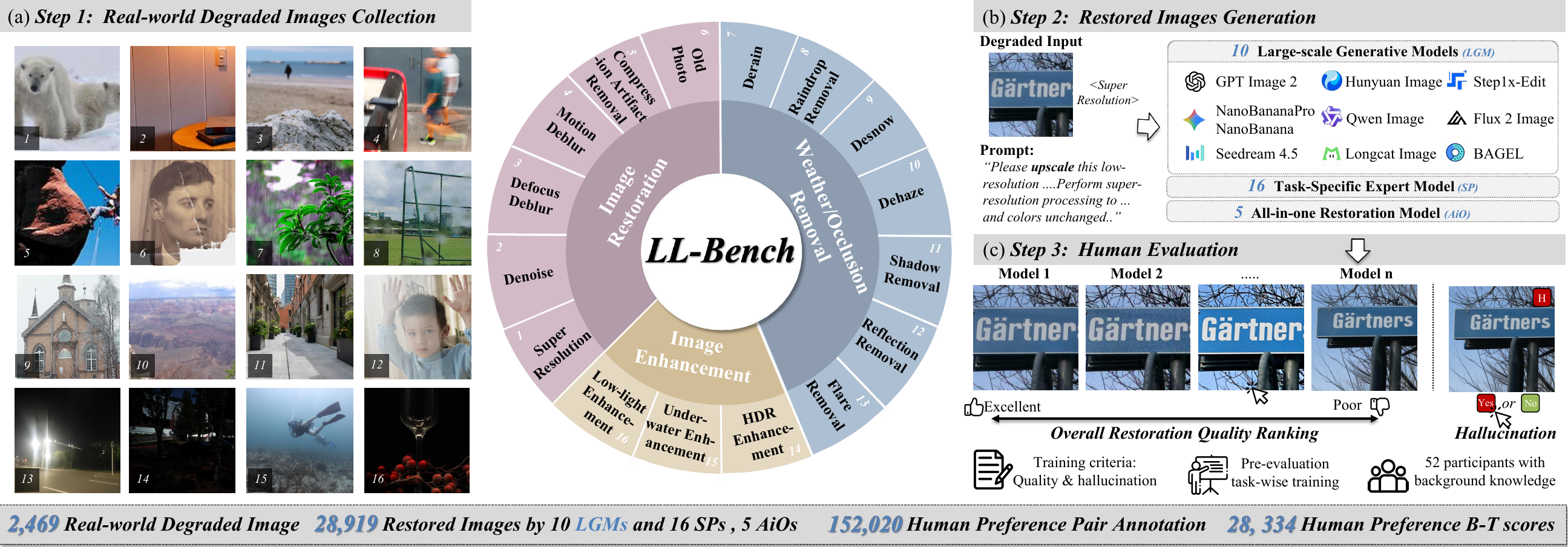}
\vspace{-1em}
\caption{\textbf{Data Construction Pipeline of LL-Bench}, a large-scale benchmark for evaluating LGMs in low-level vision tasks. (a) LL-Bench consists of 16 tasks covering diverse real-world degraded images. (b) 10 LGMs, along with 21 conventional restoration models are deployed for restored image generation. (c) Expert-level human annotations are collected to provide both quality ranking and hallucination existence detection.}
\label{fig:placeholder}
\vspace{-2em}
\end{figure}

Furthermore, leveraging human preference annotations in LL-Bench, we systematically evaluate existing FR and NR image quality assessment (IQA) metrics. The results reveal that current IQA methods exhibit weak correlation with human preferences when evaluating the quality of restored images.
To address this limitation, we propose \textbf{LL-Score}, an LMM-based quality assessor for restored images in a \textbf{one-for-all} manner across 16 low-level vision tasks.
LL-Score is optimized to predict human preferences by jointly leveraging degraded inputs and restored results, enabling it to capture perceptual failure cues and identify hallucination artifacts in restored images that are neglected by conventional objective metrics.
Extensive experiments demonstrate that LL-Score not only achieves superior alignment with human preferences compared to existing IQA metrics, but also exhibits strong potential as a reward model for generative restoration. Our key contribution can be summarized as follows:
\begin{itemize}
\item We construct LL-Bench, a large-scale benchmark designed to systematically investigate the effectiveness of large-scale generative models in low-level vision through comparison with conventional restoration models, which contains 2,469 real-world degraded images covering 16 tasks, 28,919 restored images with large-scale expert-level quality annotations.
\item We systematically analyze the performance of large-scale generative models in low-level vision by comparing them with conventional restoration methods, revealing their capability boundaries and failure modes across different tasks and degradations.
\item We propose LL-Score, an LMM-based quality assessor for both restored image quality assessment and hallucination detection. Experimental results demonstrate both its superior performance on restoration IQA tasks and its effectiveness as a reward model.
\end{itemize}

\vspace{-1em}
\section{Related Work}
\vspace{-0em}
\noindent\textbf{Low-Level Vision Tasks and Methods.}  
Low-level vision encompasses a wide range of image restoration (super-resolution~\cite{odtsr, dit4sr}, motion deblurring~\cite{evssm, hidiff}, dehazing~\cite{bilalora, dehazeXL}, deraining~\cite{unirain, csud}, etc.) and image enhancement (low-light enhancement~\cite{cidnet, gsad}, HDR enhancement~\cite{clut, llf-lut}, etc.) tasks aimed at recovering or improving degraded low-quality (LQ) inputs into high-quality (HQ) counterparts.
% The dominant technical trajectory has evolved from task-specific to all-in-one approaches, and from discriminative to generative paradigms.
% Early specialist works are predominantly discriminative, where CNN-based and later Transformer-based architectures learn direct mappings from degraded observations to clean targets under supervised objectives. These methods are often supplemented with structural priors such as geometry or depth constraints.
% Later, diffusion model-based methods have emerged as a powerful generative alternative, capable of producing rich textures and diverse outputs through a stochastic denoising process.
The dominant technical trajectory has evolved from task-specific to all-in-one approaches, and from discriminative to generative paradigms.
  Early specialist works are predominantly discriminative, where CNN-based and Transformer-based architectures learn direct mappings from degraded inputs to clean targets under supervised
  objectives~\cite{dong2014srcnn,liang2021swinir,zamir2022restormer}.
  Later, diffusion model-based methods have emerged as a powerful generative alternative, capable of producing rich textures and diverse outputs through a stochastic denoising
  process~\cite{saharia2023sr3,lin2024diffbir}.
% In parallel, all-in-one restoration methods attempt to unify multiple degradations within a single framework, reducing per-task modeling overhead, including DFPIR~\cite{dfpir}, Unirestore~\cite{unirestore}, and InstructIR~\cite{instructir}.
In parallel, all-in-one restoration methods attempt to unify multiple degradations within a single framework, reducing per-task modeling overhead~\cite{dfpir,instructir,unirestore,potlapalli2023promptir}.
However, both specialist and all-in-one methods still face substantial challenges in real-world generalization, as specialist methods are sensitive to domain shift and unified methods often require retraining or adaptation across degradation distributions.

\noindent\textbf{Large-Scale Generative Models.}  
The rapid progress of large-scale generative models has introduced a new paradigm for low-level vision.
% Built on large-scale pretraining and strong generative priors, these models show substantial zero-shot transferability across diverse image manipulation tasks, including restoration-oriented editing.
% Closed-source models such as NanoBananaPro~\cite{NanoBananaPro} and GPT-Image-2~\cite{GPT-Image-2.} have shown promising visual performance in low-level scenarios without task-specific fine-tuning, suggesting that general-purpose generative priors can reduce dependence on specialized model design.
Closed-source models such as NanoBananaPro~\cite{NanoBananaPro} and GPT-Image-2~\cite{GPT-Image-2.} have shown promising visual performance in low-level scenarios, suggesting that general-purpose generative priors can reduce dependence on specialized model design~\cite{allrounder}.
% Open-source families, including Hunyuan-Image-3~\cite{Hunyuan-Image-3.0}, Flux2~\cite{Flux2-Image-Edit}, and Qwen-Image-Edit~\cite{Qwen-Image-Edit-2511}, further expand this direction by enabling controllable adaptation and instruction-based generation.
Open-source families, including Hunyuan-Image-3~\cite{Hunyuan-Image-3.0}, Flux2-Image-Edit~\cite{Flux2-Image-Edit}, and Qwen-Image-Edit~\cite{Qwen-Image-Edit-2511}, further expand this direction by enabling controllable adaptation and instruction-based editing.
In addition, unified multimodal models such as Bagel~\cite{Bagel} also exhibit image-to-image translation capability, indicating a broader trend toward generalist visual foundation models.
% Recent studies have begun to leverage these pretrained large-scale generative models for low-level vision tasks.
% For example, ODTSR~\cite{odtsr} adopts Qwen-Image as its backbone for real-world image super-resolution, and RealRestorer adopts Step1X-Edit~\cite{Step1X-Image-Edit} as its backbone to build an all-in-one framework towards real-world image restoration.
% Although these models produce visually appealing outputs, this research line is still at an early stage due to the evaluation bottleneck. In other word, existing quality assessment methods are not sufficient to  reflect the practical effectiveness of these generative low-level models.
Although these models can produce appealing results, their open-ended generative nature makes evaluation more difficult~\cite{blau2018perception,pipal}. Existing quality assessment methods are often
  insufficient for reflecting whether such outputs are simultaneously perceptually realistic, degradation-free, and faithful to the input content.

\noindent\textbf{Quality Assessment Methods for Low-Level Vision.}  
Quality assessment in low-level vision is commonly based on full-reference (FR) and no-reference (NR) methods.
% FR methods, including pixel-level PSNR, SSIM~\cite{ssim}, perceptual feature-based LPIPS~\cite{lpips}, and DISTS~\cite{dists}, require ground truth and measure similarity between restored outputs and reference images.
FR methods, including pixel-level PSNR, SSIM~\cite{ssim}, perceptual feature-based LPIPS~\cite{lpips}, and DISTS~\cite{dists},  measure similarity between restored outputs and reference images.
% These methods remain informative for deterministic reconstruction methods, but can be misleading for generative restoration: outputs with richer perceptual details may receive lower FR scores when they deviate from the specific ground-truth realization, while oversmoothed outputs may obtain higher scores. Moreover, FR methods are inapplicable in real-world scenarios where ground truth is unavailable.
These methods remain informative for deterministic reconstruction, but can mislead generative restoration: outputs with richer details may receive lower FR scores when they deviate from the specific ground-truth realization, whereas oversmoothed outputs may score higher~\cite{blau2018perception,pipal}. Moreover, FR methods are inapplicable in real-world scenarios without ground truth.
NR methods avoid ground-truth dependence and include statistical methods such as NIQE~\cite{niqe} and BRISQUE~\cite{brisque}, as well as learning-based methods such as MANIQA~\cite{maniqa}, HyperIQA~\cite{hyperiqa}, TOPIQ~\cite{topiqa}, CLIP-IQA~\cite{clip}, and MUSIQ~\cite{musiq}. These methods estimate overall perceptual quality without reference images, but provide limited signals for content fidelity with respect to degraded inputs.
% Since both existing FR and NR methods have inherent limitations, current low-level vision studies typically report both to demonstrate the quality and fidelity. However, real-world evaluation still relies on time-consuming user studies as a substitute for FR evaluation, which severely hinders the development and deployment of practical restoration and enhancement systems. 
% This bottleneck becomes even more acute in the generative era, where most existing quality assessment methods were originally designed for discriminative restoration pipelines. 
Due to the inherent limitations of existing FR and NR methods, low-level vision studies typically report both types of metrics to characterize quality and fidelity, while relying on costly user studies for reliable subjective evaluation. This bottleneck becomes more acute in the generative era, where outputs may be perceptually plausible yet differ from the reference, or appear visually natural while being inconsistent with the input content, which motivates dedicated evaluation resources for generative low-level vision beyond conventional FR and NR measurements. As summarized in Table~\ref{tab:llbench_comparison}, existing low-level IQA benchmarks either focus on limited tasks, lack large-scale generative model outputs, or provide only overall quality annotations, making them insufficient for systematically evaluating generative models on low-level vision tasks.

    \begin{table*}[t]                                                                                                                                                      
    \centering
    \caption{Comparison of LL-Bench with existing image quality assessment benchmarks for low-level vision tasks. \textcolor{darkgreen}{\ding{51}} indicates the dataset 
  includes the corresponding feature, \textcolor{red}{\ding{55}} means not applicable or not available. LGM, SP, and AIO denote large-scale generative model, specialist
  methods, and all-in-one methods, respectively.}
    \label{tab:llbench_comparison}
    \resizebox{1\textwidth}{!}{
    \begin{tabular}{lcccccccccc}
    \toprule
    \noalign{\vspace{-1.5pt}}
    \multirow{2}{*}{Dataset} & Source & Degradation & Restored & \multirow{2}{*}{Prompt} & \multirow{2}{*}{Annotations} & \multirow{2}{*}{Tasks} &
  \multicolumn{3}{c}{Methods} & \multirow{2}{*}{Dimensions} \\
    \cmidrule(lr){8-10}
    \noalign{\vspace{-2.7pt}}
    & Images & Type & Images & & & & LGM & SP & AIO & \\
    \hline
    \noalign{\vspace{1.5pt}}
    PIPAL \cite{pipal} & 250 & Synthetic & 29K & \textcolor{red}{\ding{55}} & 1.1M Scores & 3 & \textcolor{red}{\ding{55}} & \textcolor{darkgreen}{\ding{51}} &
  \textcolor{red}{\ding{55}} & Overall Quality \\
    KADID-10k \cite{kadid} & 10,125 & Synthetic & 10K & \textcolor{red}{\ding{55}} & 303K Scores & 1 & \textcolor{red}{\ding{55}} & \textcolor{darkgreen}{\ding{51}} &
  \textcolor{red}{\ding{55}} & Overall Quality \\
    RealSRQ \cite{realsrq} & 1,620 & Syn+Real & 1.6K & \textcolor{red}{\ding{55}} & 65K Scores & 1 & \textcolor{red}{\ding{55}} & \textcolor{darkgreen}{\ding{51}} &
  \textcolor{red}{\ding{55}} & Overall Quality \\
    SRIQA-Bench \cite{sriqa} & 100 & Syn+Real & 1.1K & \textcolor{red}{\ding{55}} & 5.5K Pairs & 1 & \textcolor{red}{\ding{55}} & \textcolor{darkgreen}{\ding{51}} &
  \textcolor{red}{\ding{55}} & Overall Quality \\
    FGRestore \cite{fgrestore} & - & - & 18.4K & \textcolor{red}{\ding{55}} & 30.9K Pairs & 6 & \textcolor{red}{\ding{55}} & \textcolor{red}{\ding{55}} &
  \textcolor{red}{\ding{55}} & Quality, Alignment, Preservation \\
    Yin \textit{et al.} \cite{yin2026far} & 354 & Syn+Real & 7K & \textcolor{red}{\ding{55}} & 28K Pairs & 1 & \textcolor{darkgreen}{\ding{51}} &
  \textcolor{darkgreen}{\ding{51}} & \textcolor{red}{\ding{55}} & Detail/Sharpness/Semantic/Overall \\
    \hline
    \noalign{\vspace{1.5pt}}
    \rowcolor{gray!20}
    \textbf{LL-Bench (Ours)} & \textbf{2,469} & \textbf{Real-world} & \textbf{28.9K} & \textcolor{darkgreen}{\ding{51}} & \textbf{150K Pairs,  28.9K Scores} & \textbf{16} &
  \textcolor{darkgreen}{\ding{51}} & \textcolor{darkgreen}{\ding{51}} & \textcolor{darkgreen}{\ding{51}} & \textbf{Overall Quality, Hallucination} \\
    \noalign{\vspace{-1.5pt}}
    \bottomrule
    \end{tabular}}
    % \vspace{-10pt}
    \vspace{-2em}
  \end{table*}
\section{LL-Bench Construction}
\label{sec:llbench}

In this section, we introduce \textbf{LL-Bench}, a comprehensive human-preference benchmark for evaluating large-scale generative models on \textbf{\underline{L}}ow-\textbf{\underline{L}}evel vision tasks. LL-Bench consists of 2,469 real-world degraded images, 28,919 restored images with 152,020 preference pairs annotation and per-image hallucination labels. With its diverse task coverage and extensive human preference annotations, LL-Bench serves as a systematic resource for benchmarking foundation models in low-level vision.

\subsection{Design of Low-Level Tasks}

We select 16 representative low-level vision tasks that cover the primary challenges in real-world image restoration and enhancement. These tasks can be divided into three categories 1) \textit{Image Restoration} includes 6 tasks aimed at recovering intrinsic corrupted information: super resolution, denoise, defocus deblurring, motion deblurring, compression artifact removal, and old photo restoration. 2) \textit{Weather / Occlusion Removal} encompasses 7 tasks focused on removing external interferences: derain, raindrop removal, desnow, dehaze, shadow removal, reflection removal, and flare removal. 3) \textit{Image Enhancement} contains 3 tasks aimed at improving visual quality under challenging conditions: low-light enhancement, underwater enhancement, and HDR enhancement.
These selected tasks represent diverse degradation types commonly encountered in real-world scenarios, ranging from physical degradations to environmental factors and imaging artifacts. This diversity enables comprehensive evaluation of model capabilities across different restoration challenges.

 \subsection{Data Collection}
  \noindent\textbf{Source Images and Prompts.} Evaluating with ground-truth references enables rigorous quantitative comparison, while in-the-wild evaluation better reflects real deployment scenarios.  Hence, we collect source images from two categories: 1) \textbf{1,565} images with ground-truth references from \textbf{20} established benchmark datasets, such as RealBlur\_J \cite{realblur_j} for motion deblurring, RealSR\cite{realsr} and DRealSR \cite{drealsr} for super resolution, LoLv2 \cite{lol} for low-light enhancement, and RealDOF \cite{realdof} for defocus deblurring; 2) \textbf{904} images without ground-truth references manually collected through systematic web crawling from photography platforms such as Flickr, to better evaluate model generalization in the wild. Besides, we design task-specific instruction prompt suite that explicitly emphasize maintaining original image elements while specifying the restoration objective. At last, we obtain \textbf{2,469} source degraded images across 16 tasks and \textbf{16} task-specific prompts with an average length of 59.06 words.

\noindent\textbf{Large-Scale Generative Models.} We select {\bf 10} representative image editing models including: 1) \textbf{4} closed-source commercial systems: GPT-Image-2~\cite{GPT-Image-2.}, NanoBanana~\cite{NanoBanana},NanoBananaPro~\cite{NanoBananaPro}, and Seedream 4.5~\cite{Seedream-4.5}; 2) {\bf 6} open-source alternatives:  Bagel~\cite{Bagel}, Flux2-Image-Edit~\cite{Flux2-Image-Edit}, LongCat-Image-Edit~\cite{LongCat-Image-Edit}, Qwen-Image-Edit-Plus~\cite{Qwen-Image-Edit-2511}, Step1X-Image-Edit~\cite{Step1X-Image-Edit}, and Hunyuan-Image-3.0~\cite{Hunyuan-Image-3.0}.

\noindent\textbf{Anchor Models.} We select two categories of anchor models as the evaluation baseline: 1)\textbf{16} task-specific specialist methods: IDF \cite{idf} for denoising, ODTSR \cite{odtsr} for super resolution, EVSSM \cite{evssm} for motion deblurring, NRKNet \cite{nrknet} for defocus deblurring, Codiff \cite{codiff} for compression artifact removal, BringingOldPhotosBackToLife \cite{bringingoldphotosbacktolife} for old photo restoration, CSUD \cite{csud} for derain, DeRaindrop \cite{deraindrop} for raindrop removal, SnowMaster \cite{snowmaster} for desnow, BiLaLoRA \cite{bilalora} for dehaze, StableShadowRemoval \cite{stableshadowremoval} for shadow removal, Windowseat \cite{windowseat} for reflection removal, Deflaremamba \cite{deflaremamba} for flare removal, CIDNet \cite{cidnet} for low-light enhancement, DM-water \cite{dm-water} for underwater enhancement; 2) \textbf{5} All-in-one unified methods: FoundIR \cite{foundir}, DA-CLIP \cite{daclip}, DFPIR \cite{dfpir}, Unirestore \cite{unirestore}, and InstructIR \cite{instructir}. At last, {\bf 28,919} images were collected.
% {\it We defer more details to the Appendix (Sec. \ref{sec:1.1}.)}

\subsection{Subjective Restoration Quality Assessment}
\label{human_eval}

\noindent\textbf{Participants and Apparatus.} To ensure the comprehensiveness, fairness, and reliability of the evaluation, we recruit \textbf{52} experts with backgrounds in photography or computer vision. All annotations are performed on a 27-inch 4K Dell monitor, with the viewing distance and angle set according to the ITU-R BT.500-14 guidelines~\cite{series2012methodology}. Before annotation, all participants undergo standardized training to ensure consistent understanding of the evaluation criteria, followed by a 10-image pair qualification test to verify their eligibility. Raters were required to achieve at least 80\% agreement in pairwise preferences with the pre-labeled expert consensus.

\noindent\textbf{Main Process.} We adopt a comparative evaluation methodology where participants are presented with the degraded input image alongside all restored outputs in a side-by-side layout in trial. For each image trial, annotators are asked to evaluate the restored outputs from two perspectives: \textit{1) Overall restoration quality}: annotators rank all restored outputs from best to worst by dragging images from a randomized initial order, considering task completeness, image fidelity, and visual quality. \textit{2) Hallucination}: annotators label each restored output as either hallucination-free or  hallucinated through binary classification. Attention check is included every 50 trials, only participants who had a high agreement (SRCC $> 0.70$) with previous labeling were eligible to continue to the next session.

 \noindent\textbf{Quality Control.} In addition to the above tutorial and pre-labeling, we further implement a two-stage quality control pipeline to ensure annotation reliability. First, \textit{annotator-level filtering} removes 1 annotators whose rankings consistently show low agreement (SRCC $< 0.70$) with group consensus. Second, \textit{trial-level filtering} excludes ranking trials with insufficient inter-annotator agreement (SRCC $< 0.40$), indicating ambiguous restoration quality difficult to distinguish. We then aggregate the pairwise preferences into estimated Bradley-Terry (B-T)~\cite{bradley1952rank} score of each image.  To validate the resulting annotations, we assess annotation quality using Krippendorff's $\alpha$~\cite{hayes2007answering}, where $\alpha$ for overall preference and hallucination detection are 0.6884 and 0.7142, respectively, indicating substantial agreement. At last, we obtain \textbf{152{,}020} reliable pairwise comparisons with an average of \textbf{47.6} pairs per image, along with \textbf{28{,}334} B-T scores and per-image hallucination labels.

\section{Diagnosis of Large-Scale Generative Models}
\label{sec:diagnosis}
In this section, we analyze the low-level vision capabilities of large-scale generative models by comparing them against specialist and all-in-one restoration anchor methods. Building on the subjective evaluation, we highlight new insights and outline potential directions for future work. 

\begin{figure}

    \centering
    \includegraphics[width=1\linewidth]{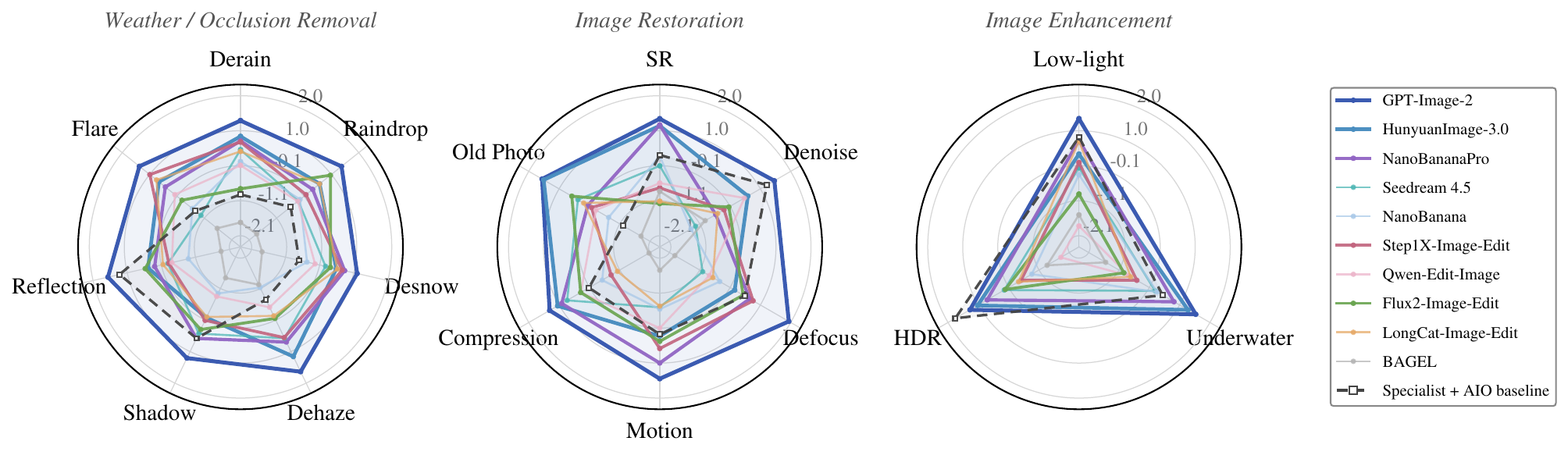}
 \vspace{-2em}
    \caption{\textbf{Averaged B-T scores of 10 large-scale generative models across 16 LL-Bench tasks.}}
    \label{fig:radar}
\vspace{-1em}
\end{figure}
\begin{conclusionbox}
\textbf{Observation 1: Large-scale Generative Models Work, but Unevenly.} 
% Large-scale generative models are not all universal restorers, capability differ by model and task.
\end{conclusionbox}
\textbf{Analysis:} 
As shown in Figure~\ref{fig:radar}, most LGMs match or surpass specialist and all-in-one baselines, demonstrating promising low-level vision capability. However, this capability varies substantially across both models and tasks.  \noindent\textbf{1) Uneven model capability.} GPT-Image-2~\cite{GPT-Image-2.} emerges as a universal restorer, with Nano Banana Pro~\cite{NanoBananaPro} and Hunyuan-Image-3.0~\cite{Hunyuan-Image-3.0} forming a strong second tier. The other models show weaknesses on particular tasks.  \noindent\textbf{2) Uneven task capability.}  (a) On \textit{weather and occlusion removal}, LGMs achieve higher and more concentrated mean B-T scores, indicating strength on semantic inpainting and understanding.  (b) On \textit{image enhancement}, B-T scores are uniformly low, indicating limited control over tone mapping and illumination. (c) \textit{Image restoration} is the most unstable category, with large inter-model variance and distinct task preferences across LGMs.

\begin{conclusionbox}
\textbf{Observation 2: Generalization Excels.} 
\end{conclusionbox}
% As shown in Figure~\ref{fig:boxplot}, conventional methods show clear performance variation between in-domain and out-of-domain datasets, such as raindrop removal, shadow removal, and motion deblurring. This variation is reflected in wider box ranges and longer diamond spans than those of LGMs. In contrast, LGMs produce consistently tighter boxes across tasks, indicating more similar performance across different datasets. We attribute this stability to the powerful open-world visual priors learned during large-scale pretraining.

\begin{figure}
    \centering
    \includegraphics[width=1\linewidth]{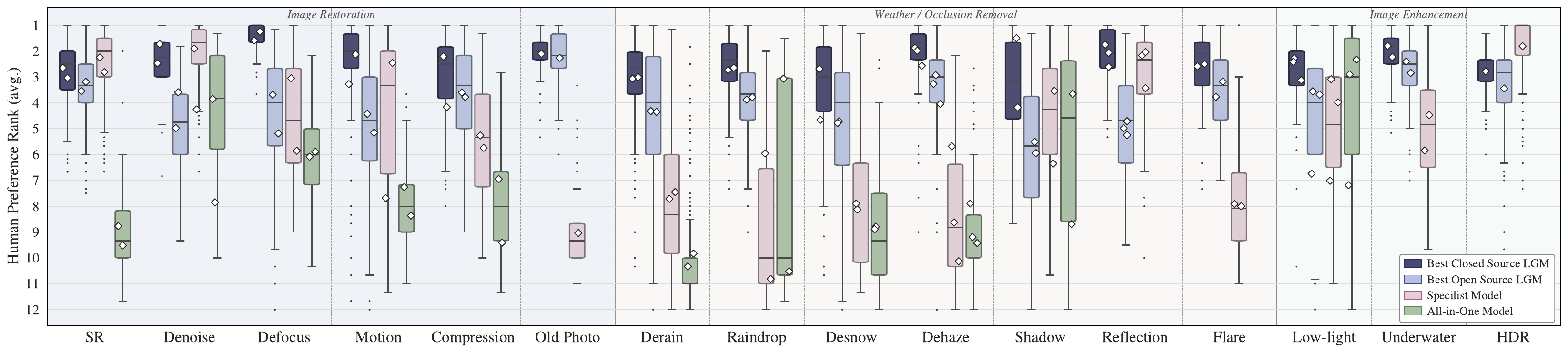}
    \vspace{-1.5em}
   \caption{\textbf{Box plots of averaged ranks across 16 LL-Bench tasks for large-scale generative models (LGMs), specialists, and all-in-one models.} Diamonds indicate the per-dataset mean rank.  }
\vspace{-1.5em}
    \label{fig:boxplot}
\end{figure}

\textbf{Analysis:} As shown in Figure~\ref{fig:boxplot}, conventional methods show clear performance variation between in-domain and out-of-domain datasets, such as raindrop removal, shadow removal, and motion deblurring. This variation appears as wider box ranges and longer diamond spans than those of LGMs. In contrast, LGMs produce tighter boxes across tasks, indicating stable performance across different datasets. We attribute this stability to open-world visual priors learned during large-scale pretraining.

\begin{conclusionbox}
\textbf{Observation 3: Failure Mode: Three Types of Hallucination.} 
\end{conclusionbox}

\begin{wrapfigure}[12]{r}{0.45\textwidth}
  \centering
  \vspace{-18pt}
  \includegraphics[width=0.45\textwidth]{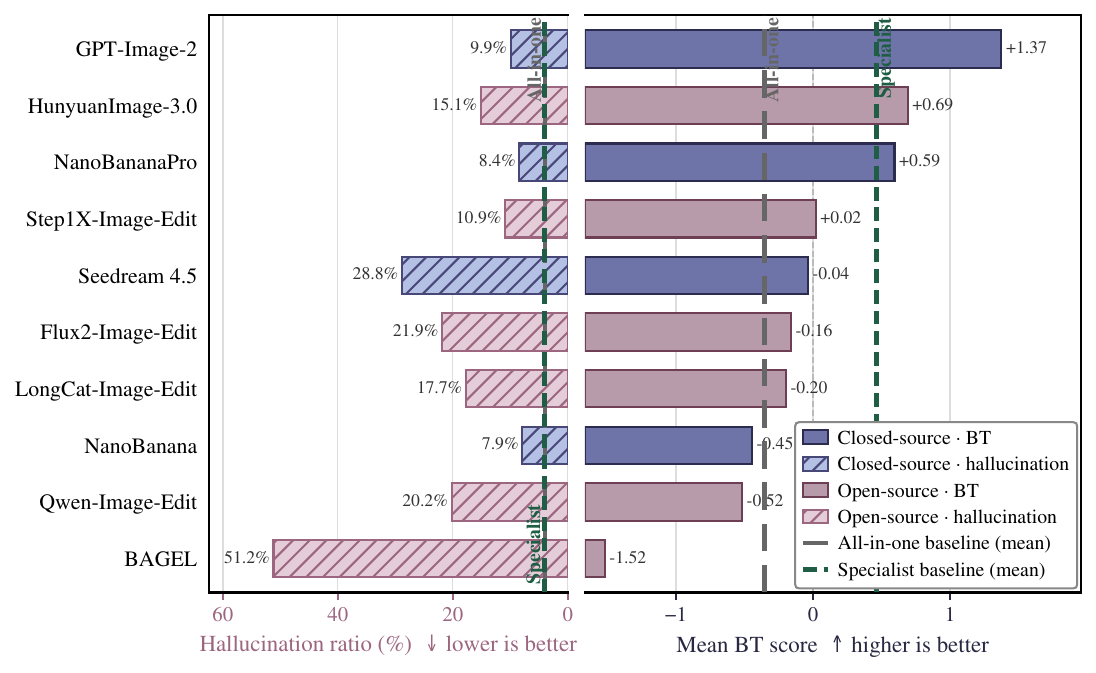}
  \vspace{-19pt}
  \caption{Comparison of averaged  hallucination ratio (\%) \textit{(left)} and B-T scores\textit{(right)}. }
  \label{fig:bar}
  \vspace{-10pt}
\end{wrapfigure}

\noindent\textbf{Analysis:} 
Figure~\ref{fig:bar} indicates although LGMs achieve strong restoration performance, all evaluated LGMs exhibit significantly higher hallucination rates than the AiO (3.98\%) and SP (4.09\%) baselines. Among LGMs, NanoBanana~\cite{NanoBanana} yields the lowest hallucination rate (7.9\%), which is still twice the average rate of conventional methods. In contrast, BAGEL~\cite{Bagel} has the highest rate at 12.8 times this average. By inspecting these failure cases, we identify three common types of hallucination: \textit{1) Color Shift:} outputs deviate from the tone, saturation, or color temperature of the input; \textit{2) Object hallucination:} small objects are removed or spuriously inserted; and \textit{3) Structural edit:} the output sometimes undergoes structural modifications, such as translation, outpainting, or cropping.

\begin{conclusionbox}
\textbf{Insight: From Generalist Editors to Faithful Restorers.} The next direction for low-level vision is faithful instruction-based editing: large-scale generative editing backbones equipped with task-specific fidelity constraints.
\end{conclusionbox}
\noindent\textbf{Analysis:} Our findings reveal that LGMs exhibit strong low-level vision capability and generalization across tasks, but fall short in fidelity and occasionally suffer from hallucination.
These findings point to instruction-based editing with explicit fidelity constraints as a promising direction.
Figure~\ref{fig:radar} and Figure~\ref{fig:bar}  provide supporting evidence: ODTSR~\cite{odtsr} and Windowseat~\cite{windowseat}, both built on fine-tuned Qwen-Image~\cite{wu2025qwen} and Qwen-Image-Editing~\cite{wu2025qwen} backbones, achieve the strongest and most stable rankings among specialist methods.

\section{Toward Preference-Aligned Quality Assessment}
\label{sec:llscore}

In this section, we present \textbf{LL-Score}, a \textbf{unified} human-preference-aligned evaluator for various low-level vision tasks. Given a restored image and its degraded input, LL-Score jointly predicts a quality score and a hallucination probability.

\subsection{Model Architecture}
\label{sec:llscore-arch}
\begin{figure}[t]
    \centering
    \includegraphics[width=1\linewidth]{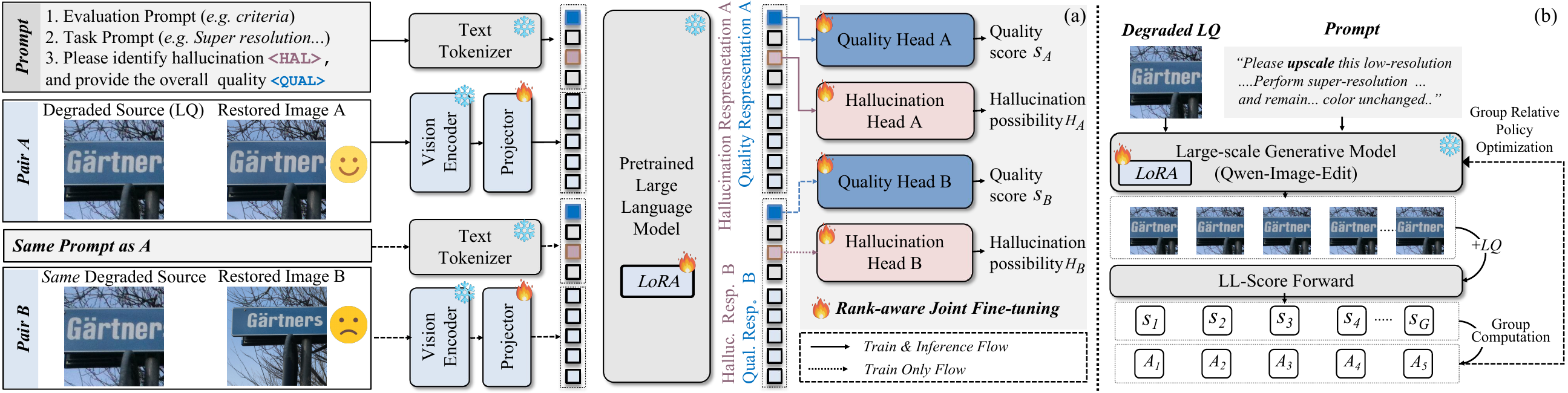}
    \vspace{-10pt}
    \caption{\textbf{(a) Overview of the LL-Score architecture. } LL-Score can evaluate overall restoration quality, and predict hallucination with the input of restored image and degraded image pairs. \textbf{(b) LL-Score as Reward Modeling.} LL-Score can act as reward signal in reinforcing LGMs.  }
    \label{fig:llscore}
\end{figure}

The overall architecture of LL-Score is depicted in Figure~\ref{fig:llscore}. The model takes a degraded source image $I_s$, the corresponding restored image $I_r$, and a prompt $T$ as inputs. Visual features are first extracted from $I_s$ and $I_r$ using the pre-trained vision encoder $E_v$ of Qwen3-VL~\cite{qwen3vl}. These features are then projected into the language embedding space through a trainable vision-language projector $P_\xi$ with parameters $\xi$, yielding the visual tokens
\begin{equation}
T_s = P_\xi(E_v(I_s)), \qquad
T_r = P_\xi(E_v(I_r)).
\end{equation}
Meanwhile, the task prompt is tokenized into text tokens $T_p$. To explicitly specify the two evaluation targets, we append two learnable special tokens, \texttt{<|QUAL|>} and \texttt{<|HAL|>}, to the prompt, indicating overall restoration quality assessment and hallucination detection, respectively. The visual tokens, text tokens, and the two special tokens are concatenated and fed into the Qwen3-VL language backbone $F_\psi$ for multi-modal feature fusion and joint reasoning. After the forward pass, the hidden states corresponding to \texttt{<|QUAL|>} and \texttt{<|HAL|>} are fed into two lightweight MLP heads, \textit{i.e.}, the quality head $Q_\omega$ and the hallucination head $H_\phi$, to produce the quality score $s_q$ and hallucination probability $p_h$. 
The whole process can be written as:
\begin{equation}
s_q, p_h =
Q_\omega\!\left(F_\psi(T_s,T_r,T_p)\right),
\quad
\sigma\!\left(H_\phi\!\left(F_\psi(T_s,T_r,T_p)\right)\right).
\end{equation}
Here, $Q_\omega$ and $H_\phi$ operate on the hidden states of \texttt{<|QUAL|>} and \texttt{<|HAL|>}, respectively. 
During training, the trainable components, including the projector, the two special-token embeddings, LoRA adapters, and the two MLP heads, are optimized jointly. 
At inference, $s_q$ is used as the LL-Score quality metric, while $p_h$ indicates the probability of hallucination.

% v2
\subsection{Training Strategy}
\label{sec:llscore-train}

As illustrated in Figure~\ref{fig:llscore}, LL-Score is trained with a \textbf{rank-aware joint fine-tuning} strategy, which combines pairwise preference learning and hallucination learning in a unified objective.

\textbf{Pairwise preference learning.}
The quality head is supervised by human pairwise preferences. According to expert consensus, we denote the preferred and less preferred restorations as $I_r^{+}$ and $I_r^{-}$, with corresponding quality scores $s_q^{+}$ and $s_q^{-}$. Each restoration also has an average human rank $r$, where a smaller value indicates stronger preference. To make the score separation reflect the confidence of human preference, we optimize a rank-aware margin-augmented Bradley-Terry loss:
\begin{equation}
\begin{aligned}
\mathcal{L}_\text{pair}
=
\frac{1}{N}
\sum_{i=1}^{N}
\log\!\Bigg[
1+\exp\!\Bigg(
s_{q,i}^{-}-s_{q,i}^{+}
+
\mathrm{clip}\!\left(
\frac{r_i^{-}-r_i^{+}}{\tau},
0,
m_{\max}
\right)
\Bigg)
\Bigg].
\end{aligned}
\label{eq:rank-bt}
\end{equation}
Here, $r_i^{+}$ and $r_i^{-}$ are the average human ranks of the preferred and less preferred restorations, respectively. The hyperparameter $\tau$ controls the sensitivity to the rank gap, and $m_{\max}$ caps the maximum margin. A larger rank gap yields a larger margin, encouraging LL-Score to assign more separated scores to clearly distinguishable restorations.

\textbf{Hallucination learning.}
In parallel, LL-Score predicts hallucination probabilities for both restored candidates. Each restored image is supervised by a binary hallucination label $y\in\{0,1\}$, where $y=1$ indicates hallucinated content. Since hallucination annotations are class-imbalanced, we train the hallucination head $H_\phi$ with binary focal loss, denoted as $\mathcal{L}_\text{hal}$. \textit{We defer detailed formulation in Appendix}~\ref{app:loss}. The final training objective combines the two losses:
\begin{equation}
\mathcal{L}
=
\mathcal{L}_\text{pair}
+
\lambda_h \mathcal{L}_\text{hal},
\label{eq:joint}
\end{equation}
where $\lambda_h$ balances preference learning and hallucination learning.

\section{Experiment}

\subsection{Implementation Details}
\label{sec:4.1}
The LLM used in LL-Score is Qwen3-VL-8B-Instruct~\cite{qwen3vl}, with a hidden channel dimension of 4096. The vision tower is frozen during training, while the LLM is adapted with LoRA and the visual merger, quality head, and hallucination head are trainable. The rank of LoRA is set to 16, and the LoRA alpha is also set to 16. LoRA dropout is set to 0.05. The hallucination head is a lightweight MLP with hidden size 1024 and bottleneck size 16. The model is trained on 4 NVIDIA A100 GPUs with DeepSpeed ZeRO-2 and flash-attn enabled. We use AdamW optimizer to train the network for 1 epoch with a per-device batch size of 1, gradient accumulation steps of 2, and learning rate $2\times10^{-6}$.
\subsection{LL-Score Evaluation}
\paragraph{Experimental Settings.}
We evaluate LL-Score on the established LL-Bench testing split, which covers 16 representative low-level vision tasks with human preference rankings and binary hallucination annotations.
Following common evaluation protocols for image quality assessment and preference prediction, we compare LL-Score with a total of 17 state-of-the-art quality assessment methods, which can be categorized into five groups:
(1) full-reference image quality assessment methods, including PSNR, SSIM~\cite{ssim}, LPIPS~\cite{lpips}, and DISTS~\cite{dists};
(2) traditional no-reference image quality assessment methods, including NIQE~\cite{niqe} and BRISQUE~\cite{brisque};
(3) deep-learning-based no-reference image quality assessment methods, including MANIQA~\cite{maniqa}, HyperIQA~\cite{hyperiqa}, TopIQA~\cite{topiqa}, CLIPIQA~\cite{clip}, and MusIQ~\cite{musiq};
(4) MLLM-based image quality assessment methods, including DeQA-Score~\cite{deqa-score}, FGResQ~\cite{fgrestore}, and Q-Align~\cite{q-align};
and (5) human-preference-aligned reward models, including HPSv3~\cite{hpsv3}, ImageReward~\cite{imagereward}, and EditReward~\cite{editreward}.
For all methods, we compute the correlation between the predicted quality scores and human preference rankings, including Spearman rank-order correlation coefficient (SRCC), and pairwise classification accuracy (Acc.).

For the hallucination prediction task, we use the same LL-Bench testing split.
Since existing low-level IQA methods do not explicitly predict hallucinations, we compare LL-Score with several representative large multimodal models in a zero-shot setting, including Qwen2-VL~\cite{Qwen2VL}, Qwen2.5-VL~\cite{Qwen2.5VL}, Qwen3-VL~\cite{qwen3vl}, DeepSeek-VL~\cite{deepseek-vl}, InternVL~\cite{chen2024internvl}, and LLaMA-3.2-Vision~\cite{llama-vision}.
We use a unified hallucination-detection prompt for all LMMs and parse their responses into binary predictions.
Accuracy, Precision, Recall, F1 score, and AUROC are adopted as the evaluation metrics.

\begin{table*}
\vspace{-0.3em}
\setlength{\belowcaptionskip}{-0.01cm}
\centering
\belowrulesep=0pt
\aboverulesep=0pt
\renewcommand\arraystretch{1.2}
\caption{Performance comparison of FR/NR metrics and our method across 16 low-level vision tasks. SRCC and Acc. are reported for each task, and SRCC$_\text{\textit{avg}}$, PLCC$_\text{\textit{avg}}$, KRCC$_\text{\textit{avg}}$ and  Acc.$_\text{\textit{avg}}$ are reported for the overall evaluation. $\spadesuit$ denotes FR IQA metrics, $\diamondsuit$ denotes traditional NR IQA metrics, $\heartsuit$ denotes deep learning-based NR IQA methods, $\circ$ denotes MLLM-based IQA models, \ding{73} denotes human preference-aligned models, respectively. The proposed method follows a \textbf{\textit{one-for-all}} fashion. The best results are highlighted in \bred{red}, and the second-best results are highlighted in \blue{blue}.}
\label{tab:metric_comparison_v3}
% \vspace{-0.8em}
   \resizebox{\linewidth}{!}{\begin{tabular}{lcccccccccccccccccc}
    \toprule[1pt]
    Dimension & \multicolumn{2}{c}{Comp. Arti. Rem.} & \multicolumn{2}{c}{Defocus Deblurring} & \multicolumn{2}{c}{Dehazing} & \multicolumn{2}{c}{Denoising} & \multicolumn{2}{c}{Deraining} & \multicolumn{2}{c}{Desnowing} & \multicolumn{2}{c}{Flare Removal} & \multicolumn{2}{c}{HDR Enhancement} & \multicolumn{2}{c}{Low-Light Enhance.} \\
  \cmidrule(lr){2-3} \cmidrule(lr){4-5} \cmidrule(lr){6-7} \cmidrule(lr){8-9} \cmidrule(lr){10-11} \cmidrule(lr){12-13} \cmidrule(lr){14-15} \cmidrule(lr){16-17} \cmidrule(lr){18-19}
 Methods / Metrics
&SRCC$\uparrow$&Acc.$\uparrow$&SRCC$\uparrow$&Acc.$\uparrow$&SRCC$\uparrow$&Acc.$\uparrow$&SRCC$\uparrow$&Acc.$\uparrow$&SRCC$\uparrow$&Acc.$\uparrow$&SRCC$\uparrow$&Acc.$\uparrow$&SRCC$\uparrow$&Acc.$\uparrow$&SRCC$\uparrow$&Acc.$\uparrow$&SRCC$\uparrow$&Acc.$\uparrow$\\
    \midrule
    $\spadesuit$PSNR & 0.4606 & 0.6742 & 0.4922 & 0.6795 & 0.3550 & 0.6228 & 0.4096 & 0.6574 & 0.3550 & 0.6307 & 0.3695 & 0.6354 & 0.1773 & 0.5633 & 0.6701 & 0.7597 & 0.7756 & 0.7923 \\
    $\spadesuit$SSIM~\cite{ssim} & 0.4938 & 0.6865 & 0.3410 & 0.6240 & 0.0957 & 0.5356 & 0.5104 & 0.6983 & 0.3500 & 0.6209 & 0.3946 & 0.6458 & -0.0403 & 0.4904 & 0.5945 & 0.7241 & 0.7124 & 0.7818 \\
    $\spadesuit$LPIPS~\cite{lpips} & 0.7917 & 0.1660 & 0.6206 & 0.2776 & 0.2951 & 0.3935 & 0.7121 & 0.2030 & 0.4521 & 0.3362 & \blue{0.6472} & 0.2617 & 0.1490 & 0.4482 & 0.7890 & 0.1898 & \blue{0.8338} & 0.1667 \\
    $\spadesuit$DISTS~\cite{dists} & 0.8282 & 0.1516 & 0.7340 & 0.2311 & 0.4795 & 0.3097 & 0.4702 & 0.2998 & \blue{0.4925} & 0.3182 & 0.6358 & 0.2632 & 0.0178 & 0.4934 & 0.8019 & 0.1774 & 0.8273 & 0.1676 \\
    \hdashline
    $\diamondsuit$NIQE~\cite{niqe} & -0.0264 & 0.5101 & -0.4564 & 0.3443 & 0.0002 & 0.4984 & -0.2436 & 0.5885 & 0.2536 & 0.5912 & 0.0487 & 0.5157 & 0.0673 & 0.5239 & -0.5243 & 0.3136 & -0.3937 & 0.3636 \\
    $\diamondsuit$BRISQUE~\cite{brisque} & 0.4479 & 0.3223 & -0.3117 & 0.3897 & 0.0545 & 0.5178 & -0.4287 & 0.6685 & -0.0922 & 0.4714 & -0.0611 & 0.4783 & 0.0018 & 0.4967 & -0.5373 & 0.3066 & -0.2821 & 0.4019 \\
    $\heartsuit$MANIQA~\cite{maniqa} & 0.7900 & 0.8269 & 0.2464 & 0.5898 & 0.2861 & 0.5992 & \blue{0.6205} & 0.7263 & 0.1546 & 0.5555 & 0.0990 & 0.5333 & 0.2729 & 0.5960 & 0.5385 & 0.6928 & 0.4615 & 0.6651 \\
    $\heartsuit$HyperIQA~\cite{hyperiqa} & 0.7394 & 0.8048 & 0.2501 & 0.5876 & 0.3126 & 0.6092 & 0.6033 & 0.7374 & 0.0653 & 0.5235 & -0.0075 & 0.4993 & 0.2041 & 0.5726 & 0.5495 & 0.6972 & 0.4940 & 0.6753 \\
    $\heartsuit$TopIQA~\cite{topiqa} & 0.7650 & 0.8214 & 0.4193 & 0.6471 & 0.2219 & 0.5726 & 0.5937 & \blue{0.7393} & -0.0952 & 0.4631 & -0.0656 & 0.4788 & 0.1111 & 0.5373 & 0.5529 & 0.6982 & 0.5036 & 0.6809 \\
    $\heartsuit$CLIPIQA~\cite{clip} & 0.7630 & 0.8177 & 0.3652 & 0.6265 & 0.3849 & 0.6418 & 0.3576 & 0.6294 & -0.1421 & 0.4483 & -0.0268 & 0.4892 & 0.2197 & 0.5726 & 0.6440 & 0.7372 & 0.5691 & 0.6999 \\
    $\heartsuit$MusIQ~\cite{musiq} & 0.7954 & 0.8306 & 0.4322 & 0.6502 & 0.2359 & 0.5785 & 0.3610 & 0.6220 & -0.2644 & 0.4051 & -0.0162 & 0.4968 & 0.1934 & 0.5687 & 0.5784 & 0.7113 & 0.5217 & 0.6852 \\
\hdashline
    $\circ$DeQA-Score~\cite{deqa-score} & 0.7833 & \blue{0.8324} & \blue{0.7954} & \blue{0.8500} & \blue{0.7076} & \blue{0.7963} & -0.0244 & 0.4886 & -0.0297 & 0.4927 & 0.4004 & 0.6353 & 0.2953 & 0.6270 & 0.8285 & 0.8543 & 0.6804 & 0.7784 \\
    $\circ$FGResQ~\cite{fgrestore} & 0.7738 & 0.8214 & 0.7276 & 0.8093 & 0.5400 & 0.7140 & 0.4870 & 0.6915 & 0.3818 & 0.6345 & 0.5331 & 0.7030 & 0.3527 & 0.6453 & 0.6378 & 0.7646 & 0.1081 & 0.5462 \\
    $\circ$Q-Align~\cite{q-align} & \blue{0.7889} & 0.8306 & 0.7696 & 0.8352 & 0.6525 & 0.7682 & 0.2690 & 0.5963 & 0.0245 & 0.5073 & 0.2668 & 0.5902 & 0.3283 & 0.6407 & \blue{0.8600} & \blue{0.8655} & 0.6655 & 0.7625 \\
    \ding{73} HPSv3~\cite{hpsv3} & 0.7841 & 0.8177 & 0.6124 & 0.7500 & 0.5648 & 0.6972 & -0.1331 & 0.4431 & -0.4491 & 0.3309 & -0.0877 & 0.4699 & -0.0405 & 0.4805 & 0.7403 & 0.7915 & 0.4736 & 0.6887 \\
    \ding{73} ImageReward~\cite{imagereward} & 0.1924 & 0.5709 & -0.0147 & 0.4963 & -0.5138 & 0.2935 & -0.3333 & 0.3706 & -0.6036 & 0.2600 & -0.4274 & 0.3177 & -0.6201 & 0.2449 & 0.3555 & 0.6278 & -0.5272 & 0.2902 \\
    \ding{73} EditReward~\cite{editreward} & 0.4270 & 0.6648 & 0.7650 & 0.8259 & 0.6442 & 0.7589 & 0.1954 & 0.5756 & 0.4173 & \blue{0.6509} & 0.6254 & \blue{0.7481} & \blue{0.4689} & \blue{0.6865} & 0.6862 & 0.7825 & 0.7892 & \blue{0.8259} \\
    \rowcolor{gray!20} \textbf{Ours} & \bred{0.8109} & \bred{0.8435} & \bred{0.8576} & \bred{0.8741} & \bred{0.7404} & \bred{0.7981} & \bred{0.6308} & \bred{0.7543} & \bred{0.7309} & \bred{0.7927} & \bred{0.7908} & \bred{0.8233} & \bred{0.8060} & \bred{0.8307} & \bred{0.9005} & \bred{0.9013} & \bred{0.8360} & \bred{0.8628} \\
    \bottomrule[1pt]
  \end{tabular}}
  % \vspace{-0.8em}
    \vspace{1pt}
   \resizebox{\linewidth}{!}{\begin{tabular}{lcccccccccccccccccc}
    \toprule[1pt]
    Dimension & \multicolumn{2}{c}{Motion Deblurring} & \multicolumn{2}{c}{Raindrop Removal} & \multicolumn{2}{c}{Reflection Removal} & \multicolumn{2}{c}{Shadow Removal} & \multicolumn{2}{c}{Super Resolution} & \multicolumn{2}{c}{Old Photo Res.} & \multicolumn{2}{c}{Underwater Enh.} & \multicolumn{4}{c}{Overall} \\
  \cmidrule(lr){2-3} \cmidrule(lr){4-5} \cmidrule(lr){6-7} \cmidrule(lr){8-9} \cmidrule(lr){10-11} \cmidrule(lr){12-13} \cmidrule(lr){14-15} \cmidrule(lr){16-19}
 Methods / Metrics
&SRCC$\uparrow$&Acc.$\uparrow$&SRCC$\uparrow$&Acc.$\uparrow$&SRCC$\uparrow$&Acc.$\uparrow$&SRCC$\uparrow$&Acc.$\uparrow$&SRCC$\uparrow$&Acc.$\uparrow$&SRCC$\uparrow$&Acc.$\uparrow$&SRCC$\uparrow$&Acc.$\uparrow$&SRCC$_\text{\textit{avg}}$$\uparrow$&Acc.$_\text{\textit{avg}}$$\uparrow$&PLCC$_\text{\textit{avg}}$$\uparrow$&KRCC$_\text{\textit{avg}}$$\uparrow$\\
    \midrule
    $\spadesuit$PSNR & 0.3581 & 0.6219 & 0.4823 & 0.6732 & 0.4629 & 0.6698 & 0.7343 & \blue{0.7664} & 0.4368 & 0.6420 & N/A & N/A & 0.6703 & 0.7490 & 0.3674 & 0.6267 & 0.3611 & 0.2506 \\
    $\spadesuit$SSIM~\cite{ssim} & 0.2361 & 0.5734 & 0.4145 & 0.6503 & 0.3561 & 0.6271 & 0.3977 & 0.6448 & 0.2873 & 0.6026 & N/A & N/A & 0.4003 & 0.6411 & 0.2725 & 0.5942 & 0.2744 & 0.1863 \\
    $\spadesuit$LPIPS~\cite{lpips} & 0.5754 & 0.2921 & 0.6706 & 0.2549 & 0.5099 & 0.3135 & \blue{0.7914} & 0.1744 & \blue{0.6878} & 0.2469 & N/A & N/A & 0.5353 & 0.3203 & 0.4074 & 0.3572 & 0.4101 & 0.2824 \\
    $\spadesuit$DISTS~\cite{dists} & 0.5979 & 0.2832 & \blue{0.7072} & 0.2277 & 0.3492 & 0.3587 & 0.6291 & 0.2723 & 0.5257 & 0.3139 & N/A & N/A & 0.3934 & 0.3641 & \blue{0.4509} & 0.3405 & \blue{0.4745} & \blue{0.3154} \\
    \hdashline
    $\diamondsuit$NIQE~\cite{niqe} & -0.5654 & 0.2974 & 0.0222 & 0.5066 & -0.1399 & 0.4536 & -0.0317 & 0.4895 & -0.4511 & 0.3519 & 0.1558 & 0.5566 & -0.2929 & 0.3986 & -0.1455 & 0.4500 & -0.1211 & -0.0990 \\
    $\diamondsuit$BRISQUE~\cite{brisque} & -0.3149 & 0.3851 & 0.0969 & 0.5304 & 0.0135 & 0.5041 & -0.1321 & 0.4535 & -0.2659 & 0.4036 & 0.1462 & 0.5475 & -0.1685 & 0.4374 & -0.1238 & 0.4576 & -0.1347 & -0.0840 \\
     $\heartsuit$MANIQA~\cite{maniqa} & 0.6885 & 0.7407 & 0.1605 & 0.5539 & 0.4194 & 0.6528 & 0.1402 & 0.5418 & 0.2112 & 0.5679 & -0.0296 & 0.4887 & 0.1244 & 0.5346 & 0.2230 & 0.5762 & 0.2247 & 0.1507 \\
    $\heartsuit$HyperIQA~\cite{hyperiqa} & 0.6662 & 0.7471 & 0.0672 & 0.5221 & 0.4465 & 0.6632 & 0.2032 & 0.5696 & 0.3527 & 0.6130 & -0.0002 & 0.5007 & 0.2457 & 0.5839 & 0.2310 & 0.5794 & 0.2434 & 0.1571 \\
    $\heartsuit$TopIQA~\cite{topiqa} & 0.7089 & 0.7565 & 0.1456 & 0.5499 & 0.3635 & 0.6285 & 0.1896 & 0.5641 & 0.2109 & 0.5741 & -0.1075 & 0.4615 & 0.2053 & 0.5649 & 0.2053 & 0.5705 & 0.2147 & 0.1394 \\
    $\heartsuit$CLIPIQA~\cite{clip} & 0.7213 & 0.7606 & 0.1907 & 0.5614 & 0.4320 & 0.6534 & 0.2836 & 0.6003 & 0.3759 & 0.6282 & 0.1781 & 0.5568 & 0.2807 & 0.5953 & 0.2694 & 0.5924 & 0.2638 & 0.1829 \\
    $\heartsuit$MusIQ~\cite{musiq} & 0.6538 & 0.7320 & 0.1678 & 0.5575 & 0.3321 & 0.6163 & 0.2797 & 0.5946 & 0.4567 & 0.6439 & 0.0719 & 0.5221 & 0.3884 & 0.6396 & 0.2446 & 0.5833 & 0.2179 & 0.1648 \\
    \hdashline
    $\circ$DeQA-Score~\cite{deqa-score} & 0.7609 & 0.8124 & 0.2423 & 0.5741 & \blue{0.6061} & \blue{0.7353} & 0.4133 & 0.6543 & 0.5189 & 0.6899 & 0.4110 & 0.6554 & 0.6711 & 0.7652 & 0.3489 & 0.7016 & 0.3380 & 0.2379 \\
   $\circ$FGResQ~\cite{fgrestore} & 0.6286 & 0.7526 & 0.6000 & 0.7373 & 0.4452 & 0.6674 & 0.2349 & 0.5905 & 0.6564 & \blue{0.7622} & \blue{0.4644} & \blue{0.6757} & \blue{0.6997} & \blue{0.7833} & 0.3530 & 0.7091 & 0.3512 & 0.2399 \\
    $\circ$Q-Align~\cite{q-align} & \blue{0.7761} & \blue{0.8227} & 0.2284 & 0.5741 & 0.5647 & 0.7104 & 0.3367 & 0.6276 & 0.5560 & 0.7185 & 0.4482 & 0.6689 & 0.6546 & 0.7652 & 0.3311 & 0.7035 & 0.3238 & 0.2267 \\
    \ding{73} HPSv3~\cite{hpsv3} & 0.6419 & 0.7464 & -0.6231 & 0.2552 & 0.0143 & 0.4910 & -0.4035 & 0.3333 & 0.3601 & 0.6399 & -0.0868 & 0.4707 & 0.5886 & 0.7427 & 0.1400 & 0.5696 & 0.0888 & 0.0959 \\
    \ding{73} ImageReward~\cite{imagereward} & -0.4598 & 0.3175 & -0.6728 & 0.2308 & -0.3766 & 0.3710 & -0.5967 & 0.2593 & 0.2058 & 0.5993 & -0.1908 & 0.4392 & 0.0276 & 0.5056 & -0.1536 & 0.3849 & -0.1459 & -0.1046 \\
    \ding{73} EditReward~\cite{editreward} & 0.5985 & 0.7340 & 0.6755 & \blue{0.7561} & 0.5088 & 0.6991 & 0.6655 & 0.7551 & 0.2384 & 0.5899 & 0.3629 & 0.6284 & 0.6281 & 0.7562 & 0.3798 & \blue{0.7124} & 0.3825 & 0.2629 \\
    \rowcolor{gray!20} \textbf{Ours} & \bred{0.7893} & \bred{0.8289} & \bred{0.7618} & \bred{0.8105} & \bred{0.6734} & \bred{0.7783} & \bred{0.7957} & \bred{0.8189} & \bred{0.7352} & \bred{0.7836} & \bred{0.8099} & \bred{0.8423} & \bred{0.7319} & \bred{0.8014} & \bred{0.6600} & \bred{0.8178} & \bred{0.6757} & \bred{0.4677} \\
    \bottomrule[1pt]
  \end{tabular}}
  \vspace{-1.2em}
\end{table*}

\begin{figure}[t]
    \centering
    \vspace{-0.5em}

    \begin{minipage}[t]{0.5\linewidth}
        \vspace{0pt}
        \centering
        \belowrulesep=0pt
        \aboverulesep=0pt
        \renewcommand\arraystretch{1.15}

        \resizebox{\linewidth}{!}{
        \begin{tabular}{lccccc}
        \toprule[1pt]
        Model & Flare & HDR & Derain & Mo. Deb. & Old Ph. \\
        \midrule
        Qwen2-VL (\textit{7B})~\cite{Qwen2VL}               & 0.4490 & 0.4286 & 0.3929 & 0.4773 & 0.5217 \\
        Qwen2.5-VL (\textit{7B})~\cite{Qwen2.5VL}           & 0.5510 & 0.5542 & \blue{0.614} & 0.5476 & \blue{0.6304} \\
        Qwen3-VL (\textit{8B})~\cite{qwen3vl}               & \blue{0.5624} & \blue{0.6371} & 0.5032 & \blue{0.5682} & 0.6087 \\
        DeepSeek-VL2-s (\textit{7B})~\cite{deepseek-vl} & 0.5306 & 0.5408 & \blue{0.6134} & 0.5456 & 0.6234 \\
        InternVL3 (\textit{8B})~\cite{chen2024internvl}     & 0.5102 & 0.5488 & 0.5357 & 0.5502 & 0.6152 \\
        LLaVA-1.5 (\textit{7B})~\cite{llama-vision}         & 0.4694 & 0.5102 & \blue{0.6071} & 0.4545 & 0.5167 \\
        \rowcolor{gray!20} \textbf{LL-Score (Ours)}         & \bred{0.6122} & \bred{0.6929} & \bred{0.8214} & \bred{0.6591} & \bred{0.7826} \\
        \bottomrule[1pt]
        \end{tabular}
        }

        \captionof{table}{
        Hallucination prediction accuracy across five low-level tasks.
        Best and second-best results are marked in \bred{red} and \blue{blue}.
        }
        \label{tab:hallucination_pred_tasks}
    \end{minipage}
    \hfill
    \begin{minipage}[t]{0.48\linewidth}
        \vspace{-1pt}
        \centering
        \includegraphics[width=\linewidth]{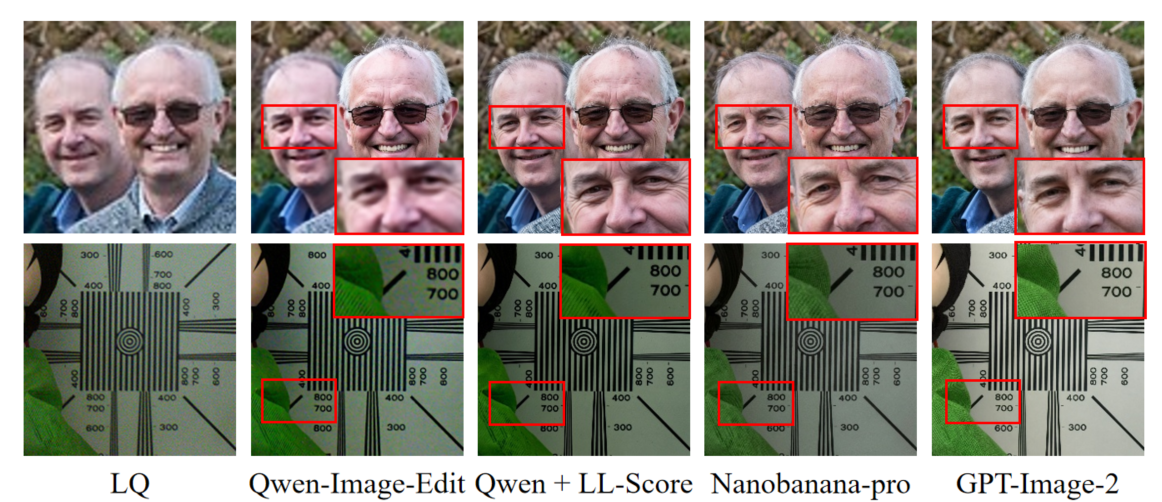}
        \caption{
        Effect of LL-Score as reward model.
        }
        \label{fig:rl_quali}
    \end{minipage}

    \vspace{-1em}
\end{figure}

\paragraph{Results on the Quality Scoring Task.}
We first evaluate LL-Score and existing quality assessment methods on LL-Bench, as shown in Table~\ref{tab:metric_comparison_v3}. 1) On image enhancement tasks, traditional FR metrics, such as PSNR and SSIM, achieve higher correlation than no-reference metrics. 2) On weather and occlusion removal tasks, FR metrics substantially outperform NR metrics, but still poorly correlated with human preference. For example, on desnowing, LPIPS (SRCC=0.49) significantly surpass MANIQA (SRCC=0.15). 3) On some general image restoration tasks, NR metrics achieve  higher correlations with human preference.  In contrast, LL-Score consistently achieves the best performance across all 16 tasks and obtains the highest overall SRCC, and pairwise accuracy. These results demonstrate that LL-Score provides a unified evaluator for diverse low-level restoration tasks and aligns better with human preferences than existing quality metrics and reward models.
\paragraph{Results on the Hallucination Prediction Task.}
  Table~\ref{tab:hallucination_pred_tasks} reports the hallucination prediction accuracy of different LMMs and LL-Score.
  We observe that zero-shot LMMs show limited and unstable performance, indicating that prompt-based detection is
  insufficient for recognizing hallucinated content in low-level restoration results.
  In contrast, LL-Score achieves the best accuracy across all evaluated tasks, demonstrating that the dedicated
  hallucination head $D_\phi$ provides more reliable hallucination supervision and complements its human-aligned quality
  scoring ability.

\subsection{LL-Score for Reward Modeling}
  \label{sec:llscore_reward_modeling}

To evaluate LL-Score as a reward model for large-scale generative editing, Qwen-Image-Edit~\cite{Qwen-Image-Edit-2511}, a strong open-source image editing model, is fine-tuned with LL-Score as a dense scalar reward. Here, we select super-resolution as a representative low-level restoration task because it requires both faithful preservation of input content and plausible recovery of high-frequency details. The policy is trained on the with ground truth super-resolution split of LL-Bench. Following Flow-GRPO~\cite{liu2025flow}, the policy samples a group of candidate edited images for each input condition using an SDE-based sampling process. LL-Score is applied to each candidate, and the group-wise rewards are normalized to compute relative advantages. The policy is then optimized to increase the likelihood of higher-scoring candidates while suppressing lower-scoring candidates. 
% Figure~\ref{fig:rl_quali} shows qualitative results on the without-GT super-resolution subset, with quantitative comparisons and more qualitative results provided in Table~\ref{tab:rl_quanti} and Figure~\ref{fig:rl_quali2} in the appendix. 
The original Qwen-Image-Edit model improves resolution to some extent, but its outputs often contain over-smoothed textures or restoration artifacts. After GRPO fine-tuning with LL-Score, the model produces sharper edges, more natural textures, and more faithful details, while better preserving the content and color distribution of the input image. These results suggest that LL-Score provides an effective optimization signal for restoration-oriented image editing, enabling fine-tuning relying on LQ images for reward computation.

\section{Conclusion}

In this work, we present \textbf{LL-Bench}, a comprehensive benchmark for evaluating large-scale generative models on low-level vision tasks. LL-Bench comprises 2,469 real-world degraded images spanning 16 tasks, 28,919 restored images generated by 10 large-scale generative models and 21 conventional restoration models, and 152,020 expert-level preference pairs together with per-image hallucination.
Besides, we provide a systematic analysis of LGMs on low-level vision tasks through comprehensive comparisons with conventional restoration models.
We further propose \textbf{LL-Score}, an MLLM-based one-for-all evaluator that jointly predicts a continuous quality score from the degraded input and identifies the hallucination probability, which can also act as a strong reward signal in reinforcing existing large-scale generative models.  We hope that LL-Bench will facilitate the development of low-level vision foundation models, offering the community a rich human-preference signal and, through LL-Score, a human-aligned evaluator that doubles as a reward.

{
  \bibliography{ref}
  \bibliographystyle{plainnat}
}

% \input{sec/08appendix}
% \newpage
% \input{Formatting_Instructions_For_NeurIPS_2026/checklist}
\end{document}